\documentclass[twocolumn]{article}
\usepackage{times}
\usepackage{caption}
\usepackage{textcomp}
\usepackage{url}
\usepackage{amsfonts,amsmath,amssymb}
\usepackage{algorithmic,algorithm}
\usepackage{fixltx2e}
\usepackage{booktabs}
\usepackage[pdftex]{graphicx}

\usepackage[varg]{txfonts}%%!!
\makeatletter%
\input{ot1txtt.fd}
\makeatother%

\title{%\bf
  Bayesian Image Restoration for Poisson Corrupted Image using a Latent Variational Method with Gaussian MRF
}

\author{Hayaru SHOUNO \\
University of Electro-Communications, Chofu, Tokyo 182--8585, Japan\\
{\tt{shouno@uec.ac.jp}}
}

\date{Dec., 07, 2014}

\def\Vec#1{\boldsymbol {#1}}

\def\avg#1{\left\langle {{#1}} \right\rangle}

\def\argmax{\mathop{\rm argmax} }

\def\ND#1#2#3{{\cal{N}}{{\left( {{#1}} \:|\: {{#2}}, {{#3}} \right)}}}

\def\Hpri{H_{\text{pri}}}

\begin{document}
\maketitle

\begin{abstract}
  We treat an image restoration problem with a Poisson noise 
  channel using a Bayesian framework. 
  The Poisson randomness might be appeared in observation of 
  low contrast object in the field of imaging. %, which means the counting of photons.
  The noise observation is often hard to treat in a theoretical analysis.
  In our formulation, 
  we interpret the observation through the Poisson noise channel as a likelihood,
  and evaluate the bound of it with a Gaussian function using a latent variable method.
  We then introduce a Gaussian Markov random field (GMRF) as the prior for the Bayesian approach,
  and derive the posterior as a Gaussian distribution.
  The latent parameters in the likelihood and the hyperparameter in the GMRF prior 
  could be treated as hidden parameters, 
  so that, 
  we propose an algorithm to infer them in the expectation maximization (EM) framework using
  loopy belief propagation(LBP).
  We confirm the ability of our algorithm in the computer simulation,
  and compare it with the results of other image restoration frameworks.
\end{abstract}

%\begin{keyword}
%  Poisson corruption process, Bayesian inference, Image restoration
%\end{keyword}

\section{Introduction}
\label{sec:introduction}
The technique of the noise reducing, 
which is called image restoration in the field of digital image processing, 
is an important in the meaning of the pre-processing.
%
%
% 20141203 add here    8<
%
In order to reduce the noise, we should focus several clues for image property.
The classical methods, such like the Gaussian or median filter methods, 
are focused on the similarity of the neighbor pixels.
%The total-variation method also focuses the similarity of 
%the nearest pixel values within the measure of $L_1$ distance\cite{Rudin92}.
In these decade, a lot of image restoration procedure were also proposed. 
%%Focusing the edge-like feature preserving, 
%%Buades \& Morel proposed non-local mean (NLM) image denoising\cite{Buades05}.
Introducing image-block similarity instead of pixel-pair similarity, 
Dabov {\it et al.} proposed a block matching method called BM3D 
that reduces the noise with less degrading edge-like features\cite{Dabov07}.
%
%%Ahron {\it et al}. proposed decomposing the image into the weighted sum of
%%several basis with introducing the clues of the sparse representation\cite{Ahron06}.
%
% >8
%

%
From the theoretical viewpoint of statistical inference,
these clues could be considered as knowledge, which is called prior, for the natural images.
Thus, it is natural to introduce Bayesian inference into the image restoration.
In the framework of Bayesian image restoration, 
additive white Gaussian noise (AWGN) was mainly discussed 
as the image corrupting process\cite{Tanaka02}\cite{Portilla03}\cite{Tanaka07}
\cite{Shouno09}\cite{Shouno11}, 
since the analytical solution could be derived explicitly for the AWGN. %% citation
However, in the real world, 
the noise corruption process often could not be described as such Gaussian observation.
For example, 
we could treat the low contrast object observation, such like night vision, 
as a Poisson noised observation, 
since the observation of photons might be expressed as a rare event.
The Poisson noised observation also appears in some kinds of 
medical imaging like positron emission tomography (PET).
%

%
% 20141203 add here 8<
%
The Poisson image restoration methods were also proposed in these decades
\cite{Rodrigues08}\cite{Decker09}\cite{Figueiredo10}\cite{Hadj13}\cite{Ono13}.
Figueiredo \& Bioucas-Dias designed the objective function as the 
the likelihood function with several penalized term, 
and
optimized the objective function with the alternating directional method of multipliers\cite{Figueiredo10}.
Ono \& Yamada proposed optimization of the similar objective function by use of hybrid steepest decent method\cite{Ono13}.
The other methods also designed the similar objective function for their applications.

These objective functions are defined by the likelihood function of Poisson observation with some penalized term.
In the Bayesian manner, regarding such penalty term as a prior, we can consider the penalized method 
as a maximization of a posteriori (MAP) method.
MAP method is a effective strategy for the image restoration, however, 
the strength balance between the prior and the likelihood is hard to determine in its framework.
On the contrary, Bayes inference could determine the strength of the penalized term naturally as the 
hyperparameter inference\cite{Tanaka02}\cite{Tanaka07}\cite{Shouno09}\cite{Shouno11}.
%
% >8
%

%
In this study, we treat a Poisson corrupted image restoration problem,
and
solve it in the manner of the Bayesian approach.
The Bayesian approach also requires both likelihood and prior probability.
We introduce the observation process as a likelihood, 
and also introduce Gaussian Markov random field (GMRF) as a prior 
after the fashion of the several works\cite{Tanaka02}\cite{Tanaka07}.

Assuming the Poisson corruption observation makes difficult to 
derive the posterior probability in analytic form, 
since the Poisson variable take discrete and non-negative value.
Thus, we introduce a latent variational approximation in the inference derivation
\cite{Palmer06}\cite{BishopBOOK06}\cite{Seeger08}\cite{Watanabe11}\cite{Sawada12}
\cite{Shouno13}\cite{Shouno14}.
In this study, 
we transform the Poisson corruption process as the corresponding Bernoulli process, 
and introduce local latent variables to approximate the observation process as the Gaussian function 
for the likelihood in the Bayesian approach\cite{Watanabe11}.
Once, we evaluate the observation likelihood as a Gaussian function, 
we can derive the posterior probability easily\cite{Shouno13}\cite{Shouno14}.
In this formulation, 
we should introduce several latent parameters to describe the observation.
In order to infer them,  we introduce a expectation maximization (EM) algorithm
\cite{Dempster77}\cite{Mackay96hyperparameters:optimize}, 
which requires an iterative inference.
Our previous work shows the preliminary results\cite{Shouno13}\cite{Shouno14} of 
this paper. 
In this paper, we refine the formulation of Bayesian inference of \cite{Shouno13}, 
and evaluate the accelerated results of \cite{Shouno14} using several images 
with comapring of other methods.

In the following, we formulate the Bayesian image framework in the section \ref{sec:Formulation} at first.
After that, we confirm the abilities of our approach with computer simulation in the
section \ref{sec:sim}.
At last, we will conclude and summarize our approach in the section \ref{sec:conclusion}.

The source code for this paper can be available from the following site: \url{https://github.com/shouno/PoissonNoiseReduction}

\section{Bayesian Formulation}
\label{sec:Formulation}
Our method is based on the Bayesian approach, so that, 
we explain both image observation process and prior probability in the following.
%
% 20141203   8<
%
Before the formulation, we define several notations.
We consider the 2-dimensional image whose size are $L_x$ and $L_y$, so that
the total number of pixels $M$ is described as $M = L_x L_y$. 
%
% >8
%

\subsection{Image Observation process}
The digital image is usually defined by the 2-dimensional pixel array.
In the observation, 
we assume the observation for each pixel is independent, 
so that we consider single pixel observation at first.
We consider each pixel has Poisson parameter $\lambda_i$ where $i$
means the position of the pixel.  
Denoting the observed pixel value as $z_i$, which means the number of
of photons for the pixel position $i$, 
we regard the observation process as the following Poisson distribution:
\begin{align}
   p( z_i \mid \lambda_i ) = \frac{(\lambda_i)^{z_i}}{z_i!} 
   \exp( -\lambda_i ).
   \label{eq:poisson1}
\end{align}
Considering the Poisson process, 
Watanabe {\it {et al.}} treat the corruption process as a Bernoulli process, 
which counts the number of on-off event in the proper time bins\cite{Watanabe11}.
Thus, we can translate eq.(\ref{eq:poisson1}) as the binomial distribution form:
\begin{align}
  p(z_i \mid \rho_i) &=\binom{K}{z_i} (\rho_i)^{z_i} (1-\rho_i)^{K-z_i},
  \label{eq:binom1}
\end{align}
where $\lambda_i = K \rho_i$, and $K$ means the upper limit of the counting.
In this formulation, we can confirm the eq.(\ref{eq:binom1}) converges to the
Poisson distribution eq.(\ref{eq:poisson1}) under the condition $K\rightarrow\infty$.

The parameter $\rho_i$ in the eq.(\ref{eq:binom1}) is a non-negative parameter, 
which is just hard to treat for us, so that 
we introduce the logit transform into the parameter $\rho_i$, that is:
\begin{align}
  x_i = \frac{1}{2} \ln \frac{\rho_{i}}{1-\rho_{i}},
\end{align}
and obtain the conditional probability for the condition $x_i$ as
\begin{align}
  p(z_i\mid x_i) &= \binom{K}{z_i} \exp( (2z_i - K) x_i - K \ln 2\cosh x_i).
  \label{eq:cond1}
\end{align}
Hence, the image corruption process can be interpreted as observing the $z_i$ under the condition of $x_i$.
%%
%% 20141203
%%
%Note that the eq.(\ref{eq:cond1}) is not a function of a observed value $z_i$ but a %parameter $x_i$.

\subsubsection{Evaluation with Local latent variables}
The term ``$\ln 2\cosh x_i$'' in the eq.(\ref{eq:cond1}) looks hard to tract for analysis.
Thus, in this study, we introduce a latent variable evaluation \cite{Palmer06}\cite{Watanabe11}.
Palmer {\it{et al.}} proposed to evaluate the lower bound of super-Gaussians
with multiplied form of the Gaussian distribution and concave parameter function\cite{Palmer06}, 
that is, any super-Gaussian, which is denoted by $p(u) = \exp( -g (u^2) )$ where $g(\cdot)$ is a concave function,
could be described as
\begin{align}
  p(u) &= \exp(-g(u^2)) \\
  &= \sup_{\xi > 0} \: \varphi(\xi) \: \ND{u}{0}{\xi^{-1}} \label{eq:lv1},\\
  \varphi(\xi) &=  \sqrt{\frac{2\pi}{\xi}} \exp\left( g^{*}\left(\frac{\xi}{2}\right) \right).
\end{align}
The function pair $g(u)$ and $g^{*}(\xi)$ is a convex conjugate relationship
which is derived from Legendre's transform
\begin{align}
  g(u) &= \inf_{\xi > 0}\xi u - g^{*}(\xi), \\
  g^{*}(\xi) &=  \inf_{u>0} \xi u - g(u).
\end{align}
Eq.(\ref{eq:lv1}) consists of the Gaussian part for $u$ and 
the non-Gaussian part described as the function $\varphi(\xi)$ where $\xi$ is a latent-parameter.
%We evaluate the bound comes from the term 
%$\ln 2\cosh x_i$ in the likelihood (\ref{eq:cond1}) 
%with substitution of $\ln 2\cosh x_i = \ln 2\cosh {\sqrt{u_i^2}} = g(u_i^2)$.
Introducing the latent parameter form, we obtain the upper bound as:
\begin{align}
  \ln 2\cosh x_i \leq \frac{\tanh \xi_i}{2\xi_i} (x_i^2 - \xi_i^2) + \ln 2\cosh \xi_i,
\end{align}
where $\xi_i$ is the latent parameter for the $i$-th pixel.
%
% 20141203  8<
%
%\begin{figure}[t]
%  \begin{center}
%    \resizebox{0.45\textwidth}{!}{
%      \includegraphics{Likelihood.pdf}
%    }
%  \end{center}
%  \caption{
%    Comparison of log-likelihood curves: The solid curve shows the true log-likelihood and 
%    both dotted and dashed are the variational curves.
%  }
%  \label{fig:Latent1}
%\end{figure}
%Fig.\ref{fig:Latent1} shows the comparison of log-likelihood curves.
%The solid curve shows the true log-likelihood, that is, 
%$2 (z-K) x - K \ln 2\cosh x + \ln \binom{K}{z}$ where $K=7$ and $z=1$.
%The dashed and dotted ones show the latent variational curves where $\xi =-0.25$ and $\xi = 1.5$ respectively.
%We can see the upper bound of each latent variational curve is true log-likelihood function.
%
% >8
%
Thus, we introduce the latent variational method into the eq.(\ref{eq:cond1}), 
we obtain the lower bound of the likelihood function:
\begin{align}
  p(z_i\mid x_i) &\geq \binom{K}{z_i} \exp( (2z_i - K) x_i - K 
  \frac{\tanh \xi_i}{2\xi_i} (x_i^2 - \xi_i^2) + \ln 2\cosh \xi_i
  )\notag\\
  &= p_{\xi_i}(z_i|x_i)
  \label{eq:likelihood1}
\end{align}

Assuming the independence for each pixel observation, 
we can easily evaluate the lower bound of whole image corruption process:
\begin{align}
  p(\Vec{z}\mid\Vec{x}) &= \prod_i p(z_i\mid x_i) \notag\\
  &\geq 
  \prod_{i}
  \binom{K}{z_i}
  \exp\left(
    -\frac{1}{2} \Vec{x}^{\text{T}} \Xi \Vec{x} + {\Vec{z}^{\prime}}^{\text{T}}\Vec{x} 
    \right)
    \notag\\
    &
    \:\:
    \exp\left(
    \frac{1}{2} \Vec{\xi}^{\text{T}} \Xi \Vec{\xi} - K \sum_{i} \ln 2 \cosh \xi_i
  \right)\notag\\
  &= p_{\Vec{\xi}}(\Vec{z} \mid \Vec{x}),
  \label{eq:cond2}
\end{align}
where $\Vec{z}^{\prime}$ means observation vector
\begin{align}
  \Vec{z}^{\prime} = \left( 2z_1 - K, \cdots, 2z_i - K, \cdots, 2z_M - K \right)^{\text{T}},
\end{align}
and $\Vec{\xi}$ means the collection of latent parameter $\{\xi_i\}$,
and matrix $\Vec{\Xi}$ means a diagonal matrix whose components are
$\{K\frac{\tanh \xi_i}{\xi_i}\}$.
Thus, we regard the lower bound of the likelihood 
$p_{\Vec{\xi}}(\Vec{z}\mid\Vec{x})$ as the observation process
which is denoted as a Gaussian form of $\Vec{x}$.

\subsection{Prior probability}
Introducing the Bayesian inference requires several prior probability for 
the image in order to compensate for the loss of information through the observation.
In this study, we assume a Gaussian Markov random field
(GMRF)\cite{Tanaka02} for the prior.
%
% 20141203    8<
%
The GMRF prior is one of the popular one in the field of image restoration, 
and it is not the state-of-art prior in the meaning of the reducing noise performance.
However, the GMRF is easy to treat in the analysis, so that we apply it in this study.
Usually, we define the GMRF as the sum of neighborhood differential square of parameters
$\sum_{(i,j)} \left( x_i - x_j\right)^2$ where $x_i$ and $x_j$ are neighborhood parameters. 
The energy function and the prior probability for the GMRF can be described as following:
\begin{align}
  \Hpri(\Vec{x}; \alpha, h) &= \frac{\alpha}{2} \sum_{(i,j)}
  (x_i - x_j)^2  + \frac{h}{2} \sum_i x_i^2 
  \label{eq:ene1} \\
  &= \frac{1}{2} \Vec{x}^{\text{t}} (\alpha\Lambda+hI) \Vec{x} 
  \\
  p(\Vec{x}|\alpha, h) &= \frac{1}{Z(\alpha,h)}
  \exp\left( 
    - \Hpri(\Vec{x}; \alpha, h)
  \right),
  \label{eq:pri1}
  \\
  Z(\alpha, h) &= \int d\Vec{x} \exp( - \Hpri(\Vec{x}; \alpha, h)) \notag\\
  &= \sqrt{|2\pi(\alpha \Lambda + hI)^{-1}|}
\end{align}
where the sum-up of $(i,j)$ means the neighborhood pixel indices, 
and the matrix $\Lambda$ and $I$ mean the adjacent and identical matrices respectively.
In the eq.(\ref{eq:ene1}), the first term means the GMRF part and 
the second means the Gaussian prior for the zero-center value for stable calculation.

\subsection{Image restoration algorithm with Posterior}
From the observation (\ref{eq:cond2}) and the prior (\ref{eq:pri1}), 
we can derive approximated posterior as
\begin{align}
  p_{\Vec{\xi}}(\Vec{x}\mid\Vec{z}, \alpha, h) 
  \propto p_{\Vec{\xi}}(\Vec{z}\mid \Vec{x}) 
  \: 
  p(\Vec{x}\mid \alpha, h),
  \label{eq:post1}
\end{align}
and the observation is evaluated with the latent-valued form, 
so that we can derive the approximated posterior as Gaussian distribution:
\begin{align}
  p_{\Vec{\xi}}(\Vec{x}\:|\:\Vec{z}, \alpha , h) &\sim \ND{\Vec{x}}{\Vec{m}}{S^{-1}}, \\
  S &= \alpha \Lambda + h I + \Xi, \\
  \Vec{m} &= S^{-1}\Vec{z}^{\prime}.
  \label{eq:post2}
\end{align}

Considering the inference parameter of $\Vec{x}$ as 
the posterior mean of the $\Vec{x}$, that is $\hat{\Vec{x}} = \avg{\Vec{x}}$,
we can obtain the inference parameter explicitly:
\begin{align}
  \avg{\Vec{x}} = \sum_{\Vec{x}}\: \Vec{x} \: p_{\Vec{\xi}}(\Vec{x}\:|\:\Vec{z}, \alpha, h) = \Vec{m}.
  \label{eq:restor1}
\end{align}

\subsection{Inference of Hyperparameters and Latent variables}
In order to obtain appropriate restoration with (\ref{eq:restor1}), 
the hyper-parameters $\alpha, h$, and the latent variables $\{\xi\}$
should be adjusted properly. 
Hereafter we introduce the notation $\theta = \{\alpha, h, \Vec{\xi}\}$ for convenience.
In order to solve,  we applied a expectation maximization (EM) algorithm for inferring 
these parameters $\Vec{\theta}$.
EM algorithm consists of two-step alternate iterations for the system that has hidden variables
\cite{Dempster77}\cite{Mackay96hyperparameters:optimize}.
Assuming the notation $t$ as the each time step, the EM algorithm could be described as
the following two-steps:
\begin{itemize}
\item {E-Step: Calculate Q-function that means the average of the likelihood function for the given parameter 
    $\theta^{(t)}$:
    \begin{align}
      {\cal{Q}}(\theta\mid\theta^{(t)}) &= \avg{\ln p(\Vec{x},\Vec{z}\mid\theta)}_{\Vec{x}\mid\theta^{(t)}}
    \end{align}
    }
\item{M-Step: Maximize the Q-function for $\theta$, and the arguments are set to the next hyper-parameters 
    $\theta^{(t+1)}$:
    \begin{align}
      \theta^{(t+1)} = \argmax_{\theta} {\cal{Q}}(\theta\mid\theta^{(t)})
    \end{align}
  }
\end{itemize}
Neglecting the constant term for the parameter $\theta$, 
we can derive the Q-function in the E-step as:
\begin{align}
  {\cal{Q}}(\theta\mid\theta^{(t)})  
  &= -\frac{1}{2} ({\Vec{m}^{(t)}}^{\text{T}} S \Vec{m}^{(t)} + \text{Tr}\:{S {S^{(t)}}^{-1}}) 
  -\frac{1}{2} \ln |\alpha\Lambda + hI| \notag\\
  & \:\: 
  +\frac{1}{2} \Vec{\xi}^{\text{T}} \Xi \Vec{\xi} - K \sum_i \ln 2\cosh \xi_i \\
  S^{(t)} &= \alpha^{(t)} \Lambda + h^{(t)} I + \Xi^{(t)} \\
  \Vec{m}^{(t)} &= {S^{(t)}}^{-1} \Vec{z}^{\prime}
\end{align}

In order to maximize the Q-function in the M-step, we solve the saddle point equations $\frac{\partial {\cal{Q}}}{\partial \alpha} = 0$, 
$\frac{\partial {\cal{Q}}}{\partial h} = 0$, and 
$\frac{\partial {\cal{Q}}}{\partial \xi_i} = 0$ for any $i$.
Thus, we obtain
\begin{align}
  \sum_i \frac{\eta_i}{\alpha \eta_i + h} &= {\Vec{m}^{(t)}}^{\text{T}} \Lambda \Vec{m}^{(t)} + \text{Tr}\: \Lambda {S^{(t)}}^{-1}, \label{eq:simul1}\\
  \sum_i \frac{1}{\alpha \eta_i + h} &= \|{\Vec{m}^{(t)}}\|^2 + \text{Tr}\: {S^{(t)}}^{-1}\label{eq:simul2},\\
  \xi_i &= \sqrt{{m_i^{(t)}}^2 + {S^{(t)}}^{-1}_{ii}},
\end{align}
where $\{\eta_i\}$ are eigenvalues of the adjacent matrix $\Lambda$ and 
${S^{(t)}}^{-1}_{ii}$ is the $(i,i)$th diagonal component of 
the matrix ${S^{(t)}}^{-1}$.

In order to obtain the exact hyper parameters $\alpha$ and $h$, 
we have to solve the eqs.(\ref{eq:simul1}) and (\ref{eq:simul2}) simultaneously, 
however, it makes increasing computational cost.
Thus, hereafter, we assume the hyperparameter $h$ is fixed and given as $h \ll \alpha$.
Then, we obtain the inference of the hyperparameter $\alpha$ as
\begin{align}
  \frac{1}{\alpha} = \frac{1}{M-1}
  \left(
  {\Vec{m}^{(t)}}^{\text{T}} \Lambda \Vec{m}^{(t)} + \text{Tr}\: \Lambda {S^{(t)}}^{-1}
  \right),
  \label{eq:alphainf}
\end{align}
since $\{\eta_i\}$, which are the eigenvalues of the adjacent matrix $\Lambda$, only has a zero component and
other components are positive values.
%Putting all of them in together, we obtain the Poisson corrupted image restoration algorithm as the Algorithm \ref{alg:EM1}.
%
%
%\begin{algorithm}[t]
%  \caption{Poisson corrupted image restoration using EM algorithm}
%  \label{alg:EM1}
%  \begin{algorithmic}[1]
%    \STATE{Set the initial hyper-parameters $\alpha^{(0)}$, $\Vec{\xi}^{(0)}$, and $h$}
%%    \STATE{Set the initial restoration image $\{x_i^{(0)}\}$}
%    \STATE{$t \leftarrow 0$}
%    \REPEAT
%      \STATE{Get mean $\Vec{m}^{(t)}$and accuracy $S^{(t)}$ for 
%        the posterior (\ref{eq:post2})
%        \begin{align}
%        S^{(t)} &= \alpha^{(t)} \Lambda + h + \Xi^{(t)} \notag\\
%        \Vec{m}^{(t)} &= {S^{(t)}}^{-1} \Vec{z}^{\prime}\notag
%        \end{align}
%      }
%%      \STATE{
%        Update parameters $\alpha$ and $\{{\xi}_i\}$
%        \begin{align}
%          \xi_i^{(t+1)} &= \sqrt{{m_i^{(t)}}^2 + {S^{(t)}}^{-1}_{ii}} \notag\\
%          \frac{1}{\alpha^{(t+1)}} &= \frac{1}{M-1}
%          \left(
%            {\Vec{m}^{(t)}}^{\text{T}} \Lambda \Vec{m}^{(t)} + \text{Tr}\: \Lambda {S^{(t)}}^{-1}
%          \right)\notag
%        \end{align}
%      }
%      \STATE{ $t \leftarrow t+1$}
%    \UNTIL{restoration image $\{m_i\}$ is converged.}
%    \STATE{$\hat{\Vec{x}} \leftarrow \Vec{m}$ as the restored image}
%    \STATE{$\hat{\lambda_i} \leftarrow K \rho(\hat{x_i})$ where $\rho(\cdot)$ is the inverse logit transform: 
%      $\rho(\hat{x_i}) = e^{\hat{x_i}} / (e^{\hat{x_i}} + e^{-\hat{x_i}})$}.
%  \end{algorithmic}
%\end{algorithm}

\subsubsection{Approximating Posterior Mean with Loopy Belief Propagation}
%In the Algorithm \ref{alg:EM1},
In the Algorithm of the previous section\cite{Shouno14},
each E-step requires the inverse of accuracy matrix 
${S^{(t)}}^{-1} = (\Xi^{(t)} + \alpha^{(t)}\Lambda + h^{(t)}I)^{-1}$ 
to calculate the parameters. 
In general, the computational cost for inverse of a matrix that 
size is $M\times M$ requires $O(M^3)$ order.
In this study, we assume the restoring image size is $M = L_x \times L_y$, 
so that, in the meaning of calculation scalability, 
the reduction of the cost is important for the application

\begin{algorithm}[t]
  \caption{Poisson corrupted image restoration using EM algorithm with LBP}
  \label{alg:EM2}
  \begin{algorithmic}[1]
    \STATE{Set the initial hyper-parameters $\alpha^{(0)}$, $\Vec{\xi}^{(0)}$, and $h$}
%    \STATE{Set the initial restoration image $x_i^{(0)}$}
    \STATE{$t \leftarrow 0$}
    \REPEAT
      \STATE{Set $\beta_i^{(t)} = K\frac{\tanh{\xi_i^{(t)}}}{\xi_i^{(t)}}$, and $y_i^{(t)} = (2 z_i-K) / \beta_i^{(t)}$.}
      \STATE{
        Carry out the LBP, where update eqs. are 
        (\ref{eq:LBPupdate1}) and (\ref{eq:LBPupdate2}),
        under the given hyper-parameters $\alpha^{(t)}, \{\beta_i^{(t)}\}$.
        }
      \STATE{
        After convergence of the LBP, solve several statistics:
        the restoration pixel values $\{m_i\}$, 
        those of variances $\{(\sigma_i)^2\}$, and
        the correlations $\{s_{ij}\}$:
        \begin{align}
          m_i &= 
          \frac{\beta_i^{(t)} y_i^{(t)} + \sum_{j\in N(i)} \gamma_{j\rightarrow i}\mu_{j\rightarrow i}}
          {\beta_i^{(t)} + h + \sum_{j\in N(i)}\gamma_{j\rightarrow i}} \\
          \sigma_i^{2} &= (\beta_i^{(t)} + h + \sum_{j\in N(i)}\gamma_{j\rightarrow i})^{-1},\\
          s_{ij} &= \frac{(\alpha^{(t)} - \gamma_{i \rightarrow j})(\alpha^{(t)} - \gamma_{j \rightarrow i})}{{\alpha^{(t)}}^3}. \label{eq:cor1}
        \end{align}
      }
      \STATE{Update the hyper-parameters:
        \begin{align}
          \xi_i^{(t+1)} &= \sqrt{{m_i}^2 + {\sigma_i}^{2}} \\
          \frac{1}{\alpha^{(t+1)}} &= \frac
          {
          \left(
            \sum_{(i,j)} (m_i - m_j)^2 + \sigma_i^2 + \sigma_j^2 -2 s_{ij}
          \right)
          }
          {M-1}.
        \end{align}
      }
      \STATE{ $t \leftarrow t+1$}
    \UNTIL{restoration image $\{m_i\}$ is converged.}
    \STATE{$\hat{\Vec{x}} \leftarrow \Vec{m}$ as the restored image}
    \STATE{$\hat{\lambda_i} \leftarrow K \rho(\hat{x_i})$ where $\rho(\cdot)$ is the inverse logit transform: 
      $\rho(\hat{x_i}) = e^{\hat{x_i}} / (e^{\hat{x_i}} + e^{-\hat{x_i}})$}.
  \end{algorithmic}
\end{algorithm}

In order to reduce the calculation cost, 
we introduce the loopy belief propagation (LBP) into the E-step in the algorithm.
In the manner of the Gaussian graphical model, 
the efficacy of the LBP were confirmed\cite{Tanaka07}\cite{Murphy:1999:LBP:2073796.2073849}\cite{Shouno14}.
Our approximated posterior, that is eq.(\ref{eq:post1}), is expressed as a kind of Gaussian form, 
so that we can apply the LBP for the restoration.
For applying LBP, 
we modify the evaluation of restoration value described as (\ref{eq:restor1}) to
the marginal posterior $p_{\Vec{\xi}}(x_i \mid\Vec{z}, \alpha, h)$ mean (MPM):
\begin{align}
  x_i^{*} = \avg{x_i}_{\text{MPM}} = \int dx_i\: x_i\: p_{\Vec{\xi}}(x_i\mid \Vec{z},\alpha, h).
\end{align}
Obtaining the marginal posterior mean, 
we apply a local message passing algorithm defined by LBP.
Hereafter, for convenience, we introduce the following notations:
\begin{align}
  \beta_i &= K \frac{\tanh \xi_i}{\xi_i}, \\
  y_i &= \frac{2 z_i - K}{\beta_i}.
\end{align} 
Then we obtain the observation likelihood (\ref{eq:likelihood1}) for $i$-th node as
\begin{align}
  p(y_i\mid x_i) \propto \exp\left(-\frac{\beta_i}{2}(y_i - x_i)^2\right).
  \label{eq:likelihood2}
\end{align}
The LBP algorithm is a kind of local message passing.
Here, we denote the message from the $j$th node to the $i$-th node as 
${\cal{M}}_{j\rightarrow i}(x_i)$. 
%
%Fig.\ref{fig:LBP1} shows the schematic diagram of the message passing.
%Here, considering the message ${\cal{M}}_{j\rightarrow i}(x_i)$, we should
%integrate the message of the $j$th connected units except $i$-th.
In each LBP iteration, this message passing is carried out for each connection.
In the GMRF case, the message can be derived as
\begin{align}
  {\cal{M}}_{j\rightarrow i}(x_i) \propto \int dx_j \:
  & p(y_j\mid x_j) 
  \exp(-\frac{\alpha}{2}(x_i - x_j)^2-\frac{h}{2}{x_j}^2) \notag\\
  & \prod_{k\in N(j)\backslash i} {\cal{M}}_{k\rightarrow j}(x_j),
  \label{eq:msg1}
\end{align}
where $N(j)$ means the collection of the connected units to the $j$th unit, 
and $N(j)\backslash i$ means the collection except $i$-th unit.
From the form of the integral in the eq.(\ref{eq:msg1}), 
we can regard the message from the $j$th node to the $i$-th node as the following Gaussian
\begin{align}
  {\cal{M}}_{j\rightarrow i}(x_i) \propto \ND{x_i}{\mu_{j\rightarrow i}}{{\gamma_{j\rightarrow i}}^{-1}}.
  \label{eq:msg2}
\end{align}
Substituting the message form eq.(\ref{eq:msg2}) into the eq.(\ref{eq:msg1}),
we can derive the message update rule as
\begin{align}
  \mu_{j\rightarrow i} &= \frac{\beta_j y_j + \sum_{k \in N(j) \backslash i} \gamma_{k\rightarrow j}\mu_{k\rightarrow j}}
  {\beta_j + \sum_{k \in N(j) \backslash i} \gamma_{k\rightarrow j} + h} 
  \label{eq:LBPupdate1}
  \\
  \frac{1}{\gamma_{j\rightarrow i}} &= \frac{1}{\alpha} + 
  \frac{1}{\beta_j + \sum_{k\in N(j)\backslash i} \gamma_{k\rightarrow j} + h}.
  \label{eq:LBPupdate2}
\end{align}
The LBP requires iterations for convergence of the message values.
After the convergence, the marginal posterior required for the EM algorithm can be evaluated as
\begin{align}
  p(x_i\mid\Vec{y},\alpha,h) &\propto p(y_i\mid x_i) \prod_{j\in N(i)} {\cal{M}}_{j\rightarrow i}(x_i), 
  \label{eq:mp1}\\
  p(x_i, x_j\mid \Vec{y}, \alpha, h) &\propto 
  p(y_i\mid x_i) p(y_j\mid x_j) \notag\\
  & \exp(-\frac{\alpha}{2} (x_i - x_j)^2 - \frac{h}{2}(x_i^2+x_j^2)) \notag\\
  & \prod_{k\in N(i)\backslash j} {\cal{M}}_{k\rightarrow i} (x_i)
  \prod_{l \in N(j)\backslash i}{\cal{M}}_{l\rightarrow j} (x_j).
  \label{eq:mp2}
%  &= \ND{x_i}
%  {\frac{\beta_i y_i + \sum_{j\in N(i)} \gamma_{j\rightarrow i} \mu_{j\rightarrow i}}{\beta_i + \sum_{j\in N(i)}\gamma_{j\rightarrow i} + h}}
%  {(\beta_i + h + \sum_{j \in N(i)} \gamma_{j\rightarrow i})^{-1}}
\end{align}
%
%\begin{figure}[t]
%  \begin{center}
%    \resizebox{0.25\textwidth}{!}{
%      \includegraphics{LoopyBeliefProp.eps}
%    }
%  \end{center}
%  \caption{
%    Schematic diagram of message passing of the LBP:
%    The LBP algorithm can be applied to infer the
%    marginalized posterior. 
%    Each circle shows the pixel, which has 4 nearest neighbors.
%    For instance, considering the message from the $j$th unit to $i$-th unit named
%    ${\cal{M}}_{j\rightarrow i}(x_i)$, the message integrate the messages from the
%    $j$th nearest neighbor except $i$-th.
%  }
%  \label{fig:LBP1}
%\end{figure}
%
Thus, the Q-function for the proposing EM algorithm is
\begin{align}
  {\cal{Q}}(\theta\mid\theta^{(t)}) &= \avg{\ln p(\Vec{x}, \Vec{y}\mid\theta)}_{\text{MPM}} \notag\\
  &= \frac{1}{2}\sum_i \ln\beta_i - \sum_i \frac{\beta_i}{2} \avg{(y_i - x_i)^2}_{\text{MPM}} \notag\\
  & + \frac{M-1}{2} \ln \alpha - \frac{\alpha}{2}\sum_{(i,j)}\avg{(x_i - x_j)^2}_{\text{MPM}},
  \label{eq:Qfunc1}
\end{align}
where $\avg{\cdot}_{\text{MPM}}$ means the average over the marginal posterior (MPM) 
denoted as eqs.(\ref{eq:mp1}) and (\ref{eq:mp2}).
Deriving the eq.(\ref{eq:Qfunc1}), 
we also assume the hyper-parameter $h$ is enough small $h/\alpha \ll 1$.

Let put them all together, the proposing LBP approximated solution is shown as the algorithm {\ref{alg:EM2}}.

\section{Computer Simulation Results}
\label{sec:sim}
\begin{figure*}[t]
  \begin{center}
    \resizebox{1\textwidth}{!}{
      \includegraphics{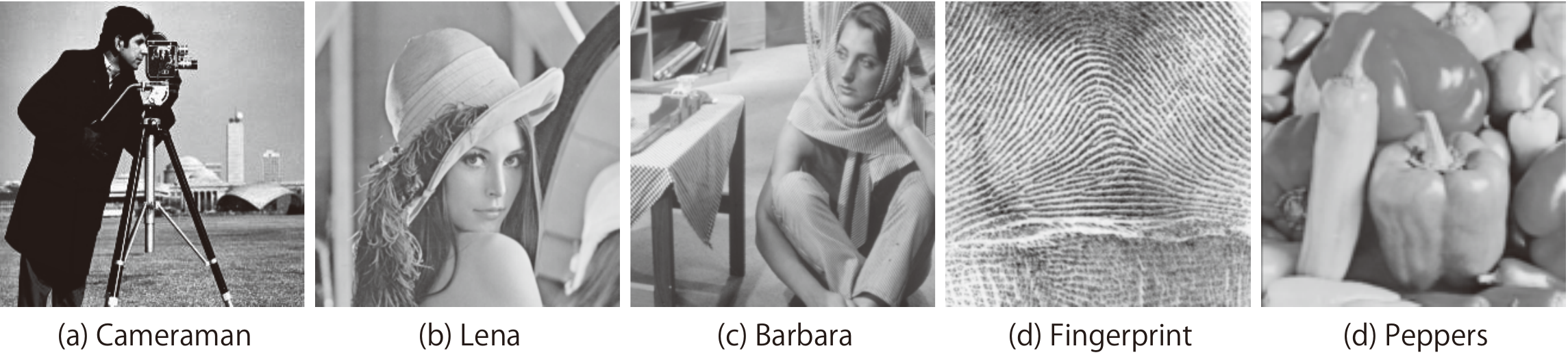}
    }
  \end{center}
  \caption{
    Images for evaluation: From (a) to  (d) show the well known evaluation image obtained from standard image database\cite{Portilla03}.
%    (e) shows an example of source image cropped from ``Peppers''. 
%    The cropping size are $L_y$ for vertical and $L_x$ for horizontal.
%    The image intensity is controlled by the parameters $\lambda_{\text{Min}}$ and $\lambda_{\text{Max}}$ shown as the 
%    eq.(\ref{eq:intensity1}).
  }
  \label{fig:imgs1}
\end{figure*}

We evaluate the restoration performance with computer simulation.
In the following, 
%at first, 
%we compare our latent variational restoration abilities 
%between the EM method solution (algorithm \ref{alg:EM1}) 
%and its LBP approximated one (algorithm \ref{alg:EM2})
%from the viewpoint of the time consumption and restored image quality.
%After that, 
we compare our latent variational restoration with LBP solution (algorithm \ref{alg:EM2})
with the conventional median filter restoration and the standard Gaussian LBP (GLBP) solution\cite{Tanaka07}.

For the evaluation, we extract several image patches from the standard images, 
called ``cameraman'', ``lena'', ``barbara'', ``fingerprint'' and ``peppers''.

We resample each image into the half-size with weak Gaussian blurring in order to increase smoothness 
since we assume observing object would be smooth.
We regard the resampled images as the observing images, 
and extract several image patches with size of $L_x \times L_y$.
The Poisson corruption process is influenced with the contrast of the observing images, 
so that, we control the maximum and minimum of the image in the simulation.
Here, we regard the patch image as $\Vec{I} = \{I_i\}$, where $i$ means the position of the pixel.
In the simulation, in order to control the contrast of the image, 
we introduce the pixel value range $(\lambda_{\text{Min}}, \lambda_{\text{Max}})$ which mean 
the minimum and the maximum values of the Poisson parameters image.
Assuming the minimum and the maximum values of the patch image as $I_{\text{Min}}$, and $I_{\text{Max}}$ respectively, 
we define the source image $\Vec{\lambda}^{*}$ of the $i$-th pixel $\lambda^{*}_i$ as a linear transform
%(see fig.\ref{fig:imgs1}(e))
:
\begin{align}
  \lambda^{*}_i = \frac{\lambda_{\text{Max}}-\lambda_{\text{Min}}}{I_{\text{Max}}-I_{\text{Min}}} (I_i - I_{\text{Min}}) + \lambda_{\text{Min}}.
  \label{eq:intensity1}
\end{align}
Thus, the difference between $\lambda_{\text{Max}}$ and $\lambda_{\text{Min}}$ becomes large, the source image becomes high contrast which means 
low noise case.
Hereafter, we fix the $\lambda_{\text{Min}} = 2$, 
and only control the parameter $\lambda_{\text{Max}}$ as the strength of the accuracy of the observation.

\subsection{Evaluation of LBP restoration abilities}
In the comparison of restoration ability, we apply the peak signal to noise ratio (PSNR).
The PSNR is defined as a kind of similarity between 
the reference image $\Vec{\lambda}^{*}$ and the test image $\Vec{\lambda}$ as:
\begin{align}
  \text{PSNR}(\Vec{\lambda}, \Vec{\lambda}^{*}) &= 10 \: \log_{10}
    \frac{  \left(\max\Vec{\lambda}^{*}-\min\Vec{\lambda}^{*}  \right)^2}
    {\text{MSE}(\Vec{\lambda}, \Vec{\lambda}^{*})}
    \label{eq:psnr1}
    ,\\
  \text{MSE}(\Vec{\lambda}, \Vec{\lambda}^{*}) &= \frac{1}{M} \sum_{i} (\lambda_i - \lambda^{*}_i)^2,
\end{align}
where $M$ means the image size $M = L_x L_y$.
f%or the algorithm \ref{alg:EM1}.
%On the contrary, in the LBP solutions, 
%the calculation cost looks insensitive to the image size. 
%Instead, the LBP solutions are affected to the corruption level, 
%when it becomes large, which means small $\lambda_{\text{Max}}$, 
%the more calculation cost is required.
%However, in the large scale image, the LBP solutions has advantage to the 
%exact solutions.

\begin{figure*}[t]
  \begin{center}
    \resizebox{1.0\textwidth}{!}{
      \includegraphics[width=1.0\textwidth]{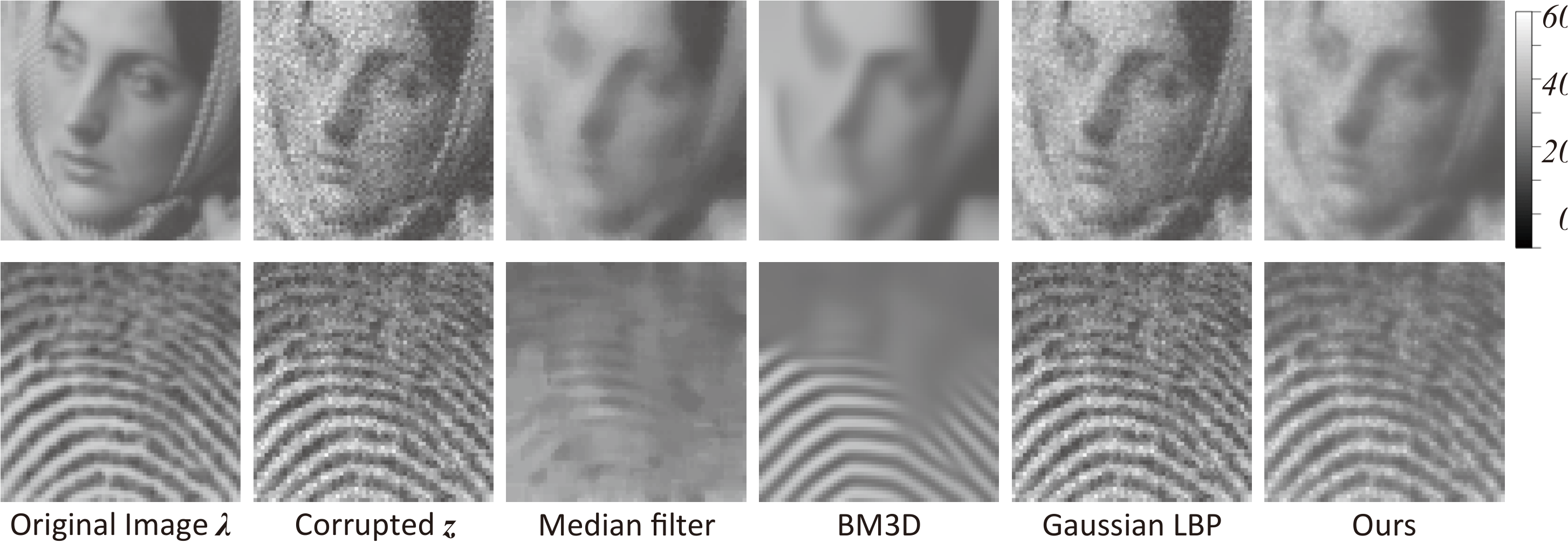}
    }
  \end{center}
  \caption{
    Comparison of restored image examples: 
    The first column shows the original image $\Vec{\lambda}$ with $L_x = L_y = 64$.
    The second shows the corrupted images through the Poisson observation
    $\Vec{z}$ 
    where the contrast parameter $\lambda_{\text{Max}} = 40$.
    The third shows the image restoration results with Median filter with $3\times 3$ size.
    The fourth shows the results of BM3D\cite{Dabov07}.
    The fourth shows the results of Gaussian LBP\cite{Tanaka07}.
    The fifth shows our latent variational method results.
  }
  \label{fig:Comp1}
\end{figure*}

%From these comparisons, 
%we show 
%the LBP approximated solution (algorithm\ref{alg:EM2}) has similar performance to the algorithm \ref{alg:EM1},
%in the meaning of the image restoration quality
%in spite of the lower computational time cost.
%Thus, we only discuss about the LBP approximated solution following section.

\subsection{Comparison with Other Image Restoration Methods}
In this section, we compare the restored image quality with the algorithm\ref{alg:EM2} 
and other image restoration methods, which are conventional median filter, BM3D method\cite{Dabov07}, and the Gaussian LBP (GLBP) solution\cite{Tanaka07}.
In these restoration methods, 
we assume restoration of the parameter $\{\lambda_i\}$
from the observed values $\{z_i\}$.
Thus, we apply the median filter, BM3D and GLBP methods to the observation $\{z_i\}$ in order to obtain the restoration result $\{\lambda_i\}$.
Especially, in the Gaussian LBP solution, the observation of the image assumes 
the pixel values $\{z_i\}$ are the result of the observation of the corresponding 
parameters $\{\lambda_i\}$ through the Gaussian channel, that is,
\begin{align}
  p(\Vec{z}\mid\Vec{\lambda}) &= \prod_{i} p(z_i\mid \lambda_i, \beta_G), \label{eq:Gauss1}\\
  p(z_i\mid\lambda_i, \beta_G) &= \sqrt{\frac{\beta_G}{2\pi}} \exp\left(-\frac{\beta_G}{2} (z_i - \lambda_i)^2\right)
\end{align}
instead of the eq.(\ref{eq:poisson1}). 
In the GLBP solution, we also adopt the GMRF as the prior
\begin{align}
  p(\Vec{\lambda} \mid \alpha_G, h_G) &= \frac{1}{Z_{\text{Gpri}}(\alpha_G, h_G)} \exp\left( - H_{\text{Gpri}}(\Vec{\lambda}\mid\alpha_G, h_G)\right), 
  \label{eq:priG}
\end{align}
instead of eq.(\ref{eq:pri1}), 
where
\begin{align}  
  H_{\text{Gpri}}(\Vec{\lambda}\mid \alpha_G, h_G) &= \frac{\alpha_{G}}{2} \sum_{(i,j)} (\lambda_i - \lambda_j)^2 + \frac{h_G}{2} \sum_i {\lambda_i}^2, \\
  Z_{\text{Gpri}}(\alpha_G, h_G) &= \sum_{\Vec{\lambda}} \exp\left( - H_{\text{Gpri}}(\Vec{\lambda}\mid\alpha_G, h_G)\right).
\end{align}
The hyperparameters $\alpha_G$, $\beta_G$, and $h_G$ are inference parameters which are solved by the EM algorithm using LBP\cite{Tanaka07}.

\begin{table}[t]
  \caption{
    Restoration quality evaluation with PSNR [dB] in fig.\ref{fig:Comp1} and median filters.
  }
  \label{tbl:Comp1}
  \begin{center}
  \begin{tabular}{|l|c|c|c|c|c|}
    \hline
                   & Corrupted & Median & BM3D  & GLBP  & Ours        \\
    \hline\hline                       
    cameraman      & 20.24     & 16.76  & 18.75 & 20.72 & \bf{21.92} \\
    \hline                             
    barbara        & 19.18     & 22.10  & 21.91 & 22.48 & \bf{23.89} \\    
    \hline                             
    fingerprint    & 19.47     & 13.83  & 18.75 & 19.86 & \bf{21.85} \\    
    \hline                             
    peppers        & 18.44     & 21.38  & 21.63 & 21.03 & \bf{22.81} \\    
    \hline
  \end{tabular}
  \end{center}
\end{table}

Fig.\ref{fig:Comp1} shows the comparison of result examples.
In the figure, we show several cropped images with $L_x = L_y = 64$ from 
``cameraman'', ``barbara'', ``fingerprint'', and ``peppers''.
In the evaluation, we fix the contrast control parameter $\lambda_{\text{Max}} = 40$ 
that controls the noise strength in the Poisson corruption.
The first column shows the original images, 
and the second one shows the Poisson corrupted images.
The third shows the conventional median filter restoration results with the size of
$3\times 3$ filter. % citation for EBimage
%
% 20141203
%
The fourth shows the restoration results with BM3D method\cite{Dabov07}.
The BM3D method is one of the state-of-art method for the AWGN, however, 
in this case, the variance of noise is not uniform in the image so that
the estimation of thresholding parameter looks insufficient.
The fifth shows the restoration results with GLBP method, and 
the sixth shows the one with our LBP method.
We can see our result images are more smooth than those of the GLBP results.
Table \ref{tbl:Comp1} shows the PSNR evaluations for each original image.
Our latent variational method shows better restoration results rather than
those of the GLBP solutions.
We also restoration results with conventional median filter with $3\times 3$ 
in the table \ref{tbl:Comp1}. 
The median filter restoration make the image too much smooth, 
so that the PSNR evaluation tends to be the small value.

In order to compare the quantitative restoration evaluations, 
we introduce the following improvement of PSNR (ISNR) index
for two type of restoration results $\Vec{\lambda}_1$ and $\Vec{\lambda}_2$,
\begin{align}
  \text{ISNR}( \Vec{\lambda}_1, \Vec{\lambda}_2; \Vec{\lambda}^{*}) 
  &= 
  \text{PSNR}(\Vec{\lambda}_2, \Vec{\lambda}^{*}) - 
  \text{PSNR}(\Vec{\lambda}_1, \Vec{\lambda}^{*})
\notag\\
  &= 10 \log 
  \frac{\text{MSE}(\Vec{\lambda}_1,\Vec{\lambda}^{*})}
  {\text{MSE}(\Vec{\lambda}_2, \Vec{\lambda}^{*})},
\end{align}
where $\Vec{\lambda}^{*}$ means the ground-truth source image.
This index shows the improvement of the $\Vec{\lambda}_2$ against 
the $\Vec{\lambda}_1$ in the meaning of PSNR.
The positive index shows the improvement of the method $\Vec{\lambda}_2$ 
from $\Vec{\lambda}_1$.

We evaluate ISNR between the noised image $\Vec{z}$ and 
our results $\Vec{\lambda}_{\text{ours}}$ that is $\text{ISNR}(\Vec{z}, \Vec{\lambda}_{\text{ours}})$, 
and also evaluate the one with other restoration results, that is 
median filter result $\Vec{\lambda}_{\text{Med}}$,
BM3D restoration result $\Vec{\lambda}_{\text{BM3D}}$,
and GLBP result $\Vec{\lambda}_{\text{GLBP}}$,
with our result, that is 
$\text{ISNR}(\Vec{\lambda}_{\text{Med}}, \Vec{\lambda}_{\text{ours}})$
$\text{ISNR}(\Vec{\lambda}_{\text{BM3D}}, \Vec{\lambda}_{\text{ours}})$
$\text{ISNR}(\Vec{\lambda}_{\text{GLBP}}, \Vec{\lambda}_{\text{ours}})$ respectively.
In the image preparation, 
we crop $10$ patch images with the size of $L_x = L_y = 64$ from random 
locations of each original image shown in fig.\ref{fig:imgs1}.
Thus, the total number for the evaluation images is $50$ image patches.
In the evaluation, we apply several contrast parameter cases, that is
$\lambda_{\text{Max}} = \{20, 40, 60, 80, 100, 120, 140, 160\}$.

\begin{figure}[t]
  \begin{center}
    \resizebox{0.45\textwidth}{!}{
      \includegraphics{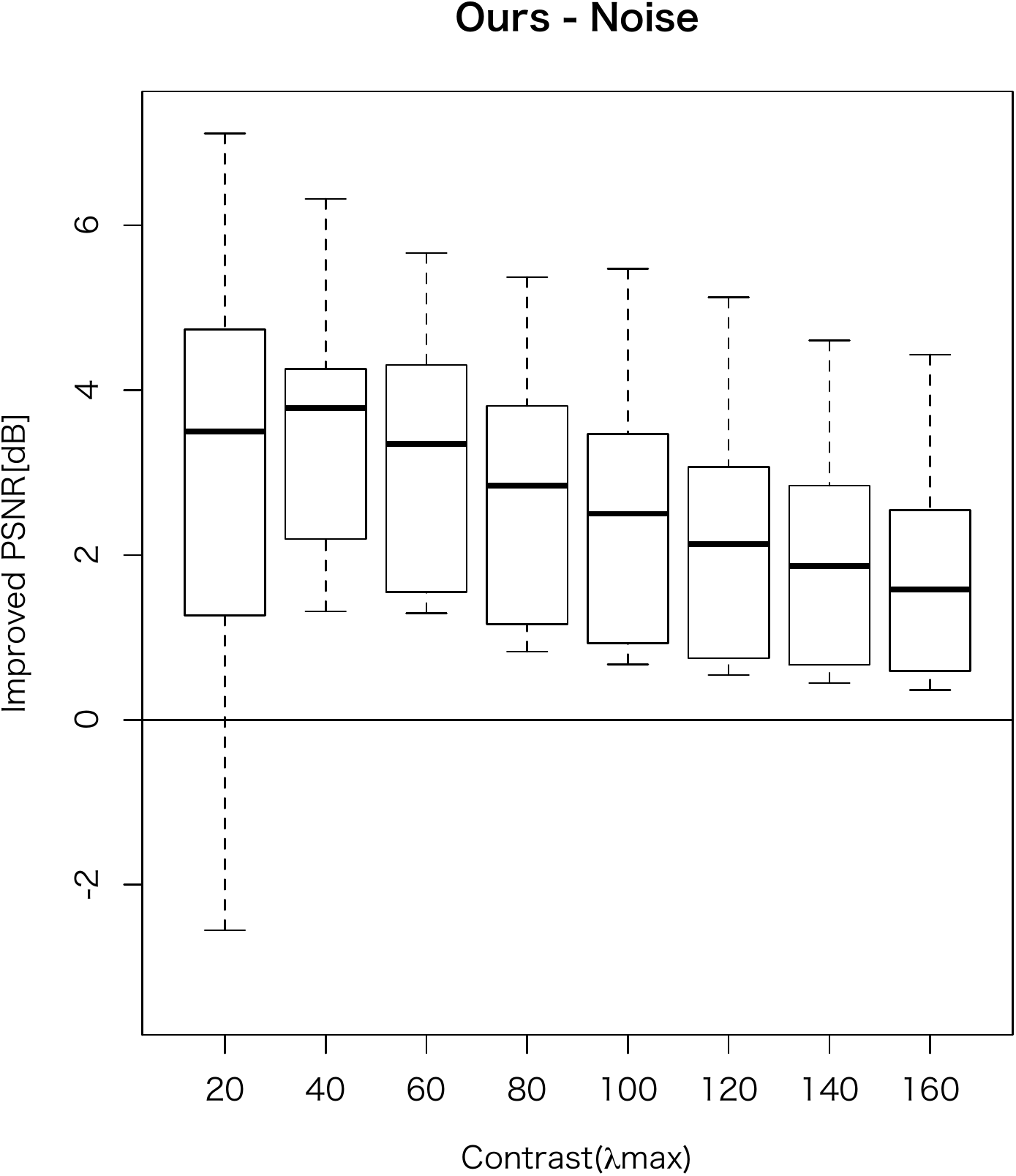}
    }
  \end{center}
  \caption{
    PSNR improvement of our results $\Vec{\lambda}_{\text{ours}}$ 
    from the observed image $\Vec{z}$.
    The horizontal axis shows the contrast parameter $\lambda_{\text{Max}}$.
    The vertical one shows the ISNR value whose range is $[-4, 7]$ [dB].
    The box-plot shows the medians with quantiles. 
  }
  \label{fig:ISNRLN}
\end{figure}

\begin{figure}[t]
  \begin{center}
    \resizebox{0.45\textwidth}{!}{
      \includegraphics{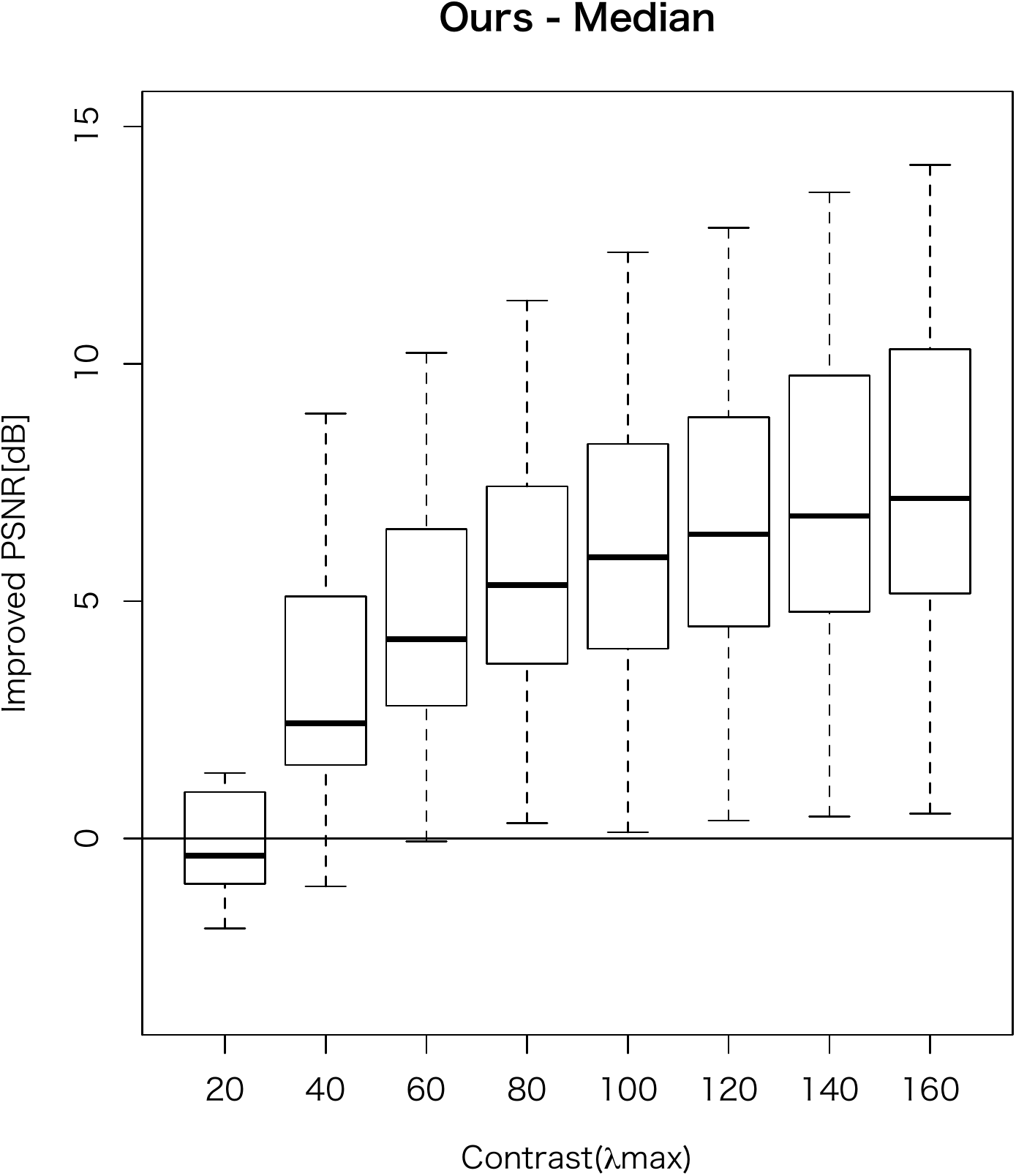}
    }
  \end{center}
  \caption{
    PSNR improvement of our results $\Vec{\lambda}_{\text{ours}}$ 
    from the median filter results $\Vec{\lambda}_{\text{Med}}$. 
    The range of vertical axis is $[-4,15]$ [dB].
  }
  \label{fig:ISNRL3}
\end{figure}

\begin{figure}[t]
  \begin{center}
    \resizebox{0.45\textwidth}{!}{
      \includegraphics{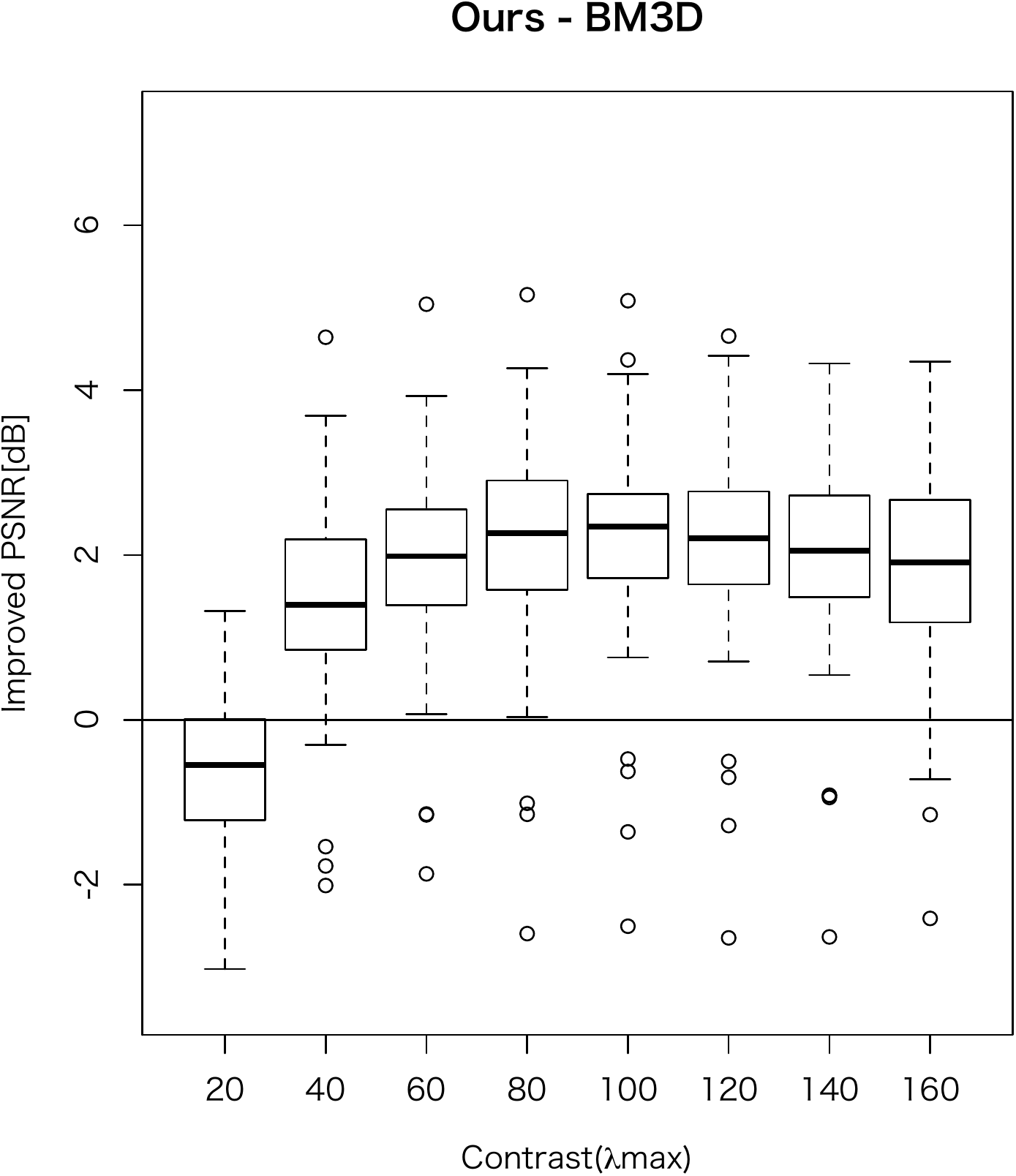}
    }
  \end{center}
  \caption{
    PSNR improvement of our results $\Vec{\lambda}_{\text{ours}}$ 
    from the BM3D solutions $\Vec{\lambda}_{\text{BM3D}}$.
    The range of vertical axis is $[-4,7]$ [dB].
    The white circles show the outliers.
  }
  \label{fig:ISNRLB}
\end{figure}

\begin{figure}[t]
  \begin{center}
    \resizebox{0.45\textwidth}{!}{
      \includegraphics{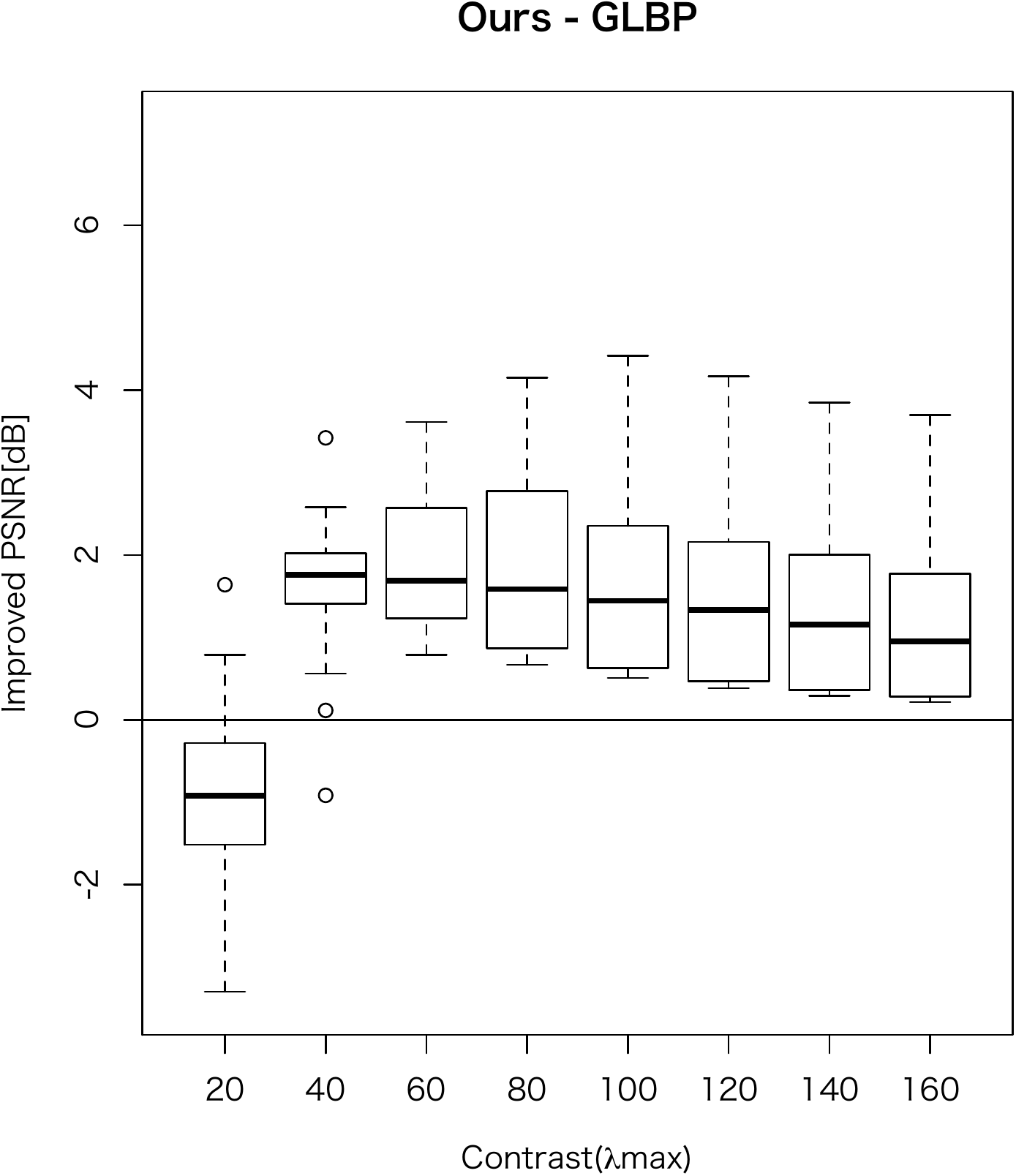}
    }
  \end{center}
  \caption{
    PSNR improvement of our results $\Vec{\lambda}_{\text{ours}}$ 
    from the GLBP solutions $\Vec{\lambda}_{\text{GLBP}}$.
    The range of vertical axis is $[-4,7]$ [dB].
    The white circles show the outliers.
  }
  \label{fig:ISNRLG}
\end{figure}

Fig.\ref{fig:ISNRLN} shows the improvement of our results for the corrupted images
$\Vec{z}$. 
The contrast parameter $\lambda_{\text{Max}}$,
which denotes the horizontal value of the figure, 
controls the noise strength of the Poisson noise.
In the figure, the boxplot shows the median, which are described as the thick line in the box, with quantiles for each contrast levels. 
we obtain $2\sim 3$ [dB] improvement from the corrupted 
image in the meaning of the median.
From the range of the control parameter $\lambda_{\text{Max}} \ge 40$, 
our method shows good performance, however, the result shows large quantile variance 
in the $\lambda_{\text{Max}}=20$.
The reason of the large variance comes from the input image property.
The low improvement results only come from the ``fingerprint'' input image,
so that, high spatial frequency with low contrast image might prevent restoration.

We also show the comparison results with applying conventional median filter.
We apply $3 \times 3$ median filter for the noised image $\Vec{z}$, and
evaluate the PSNR with the original Poisson parameters $\Vec{\lambda}$.
Fig.\ref{fig:ISNRL3} shows the results. 
The horizontal axis shows the same range of the fig.\ref{fig:ISNRLN}, 
and the vertical one shows the ISNR with the range of $[-4, 15]$ [dB].
In many cases, applying the median filter, 
the restored image becomes too much smooth, 
so that, the PSNR evaluation becomes worse.

Fig.\ref{fig:ISNRLB} shows the improvement of our results for the BM3D restoration\cite{Dabov07}.
The horizontal axis is identical to the fig.\ref{fig:ISNRLN} and \ref{fig:ISNRL3}, 
and the vertical shows the ISNR with range of $[-4, 7]$ [dB].
In the figure, the white circles show the outliers which differ twice the standard deviation or 
more from each average.
In the $\lambda_{\text{Max}} = 20$ case, the BM3D restoration looks better than that of ours.
However, in other cases, our method shows better performance.

Fig.\ref{fig:ISNRLG} shows the improvement of our results for the GLBP restoration.
This result also shows our method shows better results than that of those of
the GLBP in the $\lambda_{\text{Max}} \ge 40$.
In the range of $\lambda_{\text{Max}} \ge 40$, 
we obtain $1\sim 2$ [dB] improvement from the GLBP solution in the meaning of the
median.
On the contrary, in the $\lambda_{\text{Max} = 20}$ case, 
almost all the restoration results of the GLBP are better than those of ours.
We consider the results comes from the accuracy of the hyperparameter inference.
In the GLBP method, the noise variance, 
which is the inverse of the accuracy parameter $\beta_G$ in eq.(\ref{eq:Gauss1}),
is a single parameter.
In the low $\lambda_{\text{Max}}$ case, 
the variances of each observation $z_i$ might be described as a single value.
However, the differences of variances of the observation values $z_i$ 
would be large when the contrast parameter $\lambda_{\text{Max}}$ becomes large.
As the result, 
the Gaussian model that is eq.(\ref{eq:Gauss1}) could not describe the
observation value $z_i$ within a single hyperparameter $\beta_G$.
As the result, in many cases of the large $\lambda_{\text{Max}}$,
the inference value of the $\beta_G$ becomes large, which make low 
efficacy of the prior.

\section{Conclusion}
\label{sec:conclusion}
In this study, 
we propose an image restoration method for Poisson corrupted image.
Introducing the latent variable method, 
we derive the corruption process,
which denote the likelihood function, as a Gaussian form.
Using Bayesian image restoration framework, 
we derive the posterior probability for the restoration with 
introducing GMRF as a prior.
The posterior includes several hyperparameters $\alpha$, $h$,
and latent variables $\{\xi_i\}$.
In order to solve the restoration problem with determining these parameters, 
%we construct an algorithm 
%as the algorithm \ref{alg:EM1} in the manner of the EM method.
we construct an algorithm in the manner of the EM method.

Thus, our algorithm could determine all the parameters from the observed data $\Vec{z}$.
%The algorithm \ref{alg:EM1} requires $O(M^3)$ computational cost.
The original EM algorithm requires $O(M^3)$ computational cost.
Hence, in order to accelerate the algorithm,
we approximate the posterior mean as the marginal posterior mean,
and derive LBP method for hyperparameter inference that is described as
the algorithm \ref{alg:EM2}.
We introduce the two-body marginalized posterior described as eq.(\ref{eq:mp2})
in order to infer the correlation between connected two units denoted as $s_{ij}$ in the eq.(\ref{eq:cor1}).
%
%Usually, the LBP only consider the single-body marginal posterior described as 
%eq.(\ref{eq:mp1}). 
%Only considering the single-body marginal posterior, 
%the correlation of connected two units, which is denoted as $s_{ij}$ 
%in the eq.(\ref{eq:cor1}), becomes $0$.
%This means same effect to the naive mean field approximation.
Without the two-body interactions, the inference ignores the correlations, 
that is, $s_{ij} = 0$ for any indices.
It is identical to the naive mean field approximation.
The naive mean field approximation, 
which only applies the single-body marginal posterior, 
occurs the underestimation of the hyperparameter of $\alpha$.
%Avoiding the underestimation, we introduce the two-body marginal posterior 
%described as eq.(\ref{eq:mp2}) in the hyper-parameter inference.
The correlation $s_{ij}$ update rule is derived as the eq.(\ref{eq:cor1}), 
which only requires the local message, so that
the cost for the inference does not increase so much.
Solving exact correlation between two units requires considering not only 
the connected bodies effect but also all the other bodies effect.
This is the reason for the requiring the inverse of the accuracy matrix
in the EM algorithm.
We only consider the two-bodies effect, however, the hyper-parameter inference
looks work well, and the restoration performance becomes same or more than 
the that of the exact solution in the previous work.
Hence, we propose the LBP method is a good approximation for our 
Poisson corrupted image restoration framework.

%In the computer simulation, at first, we compare the restoration abilities of 
%the algorithms \ref{alg:EM1} and \ref{alg:EM2}.
%The algorithm \ref{alg:EM2} is a LBP approximated version of the latent variable method 
%described as the algorithm \ref{alg:EM1}.
%From the viewpoint of the hyperparameter inference 
%and the quality of the restored image, we confirm these two algorithms have similar abilities, 
%however,
%the computational cost of the algorithm \ref{alg:EM2} is lower than that of the algorithm \ref{alg:EM1}.

Then, we compare our algorithm with other methods, that is, median filter, BM3D and, GLBP restorations\cite{Dabov07}\cite{Tanaka07}.
The BM3D and GLBP method regards the obtained variables $\Vec{z}$ are observed 
from the Gaussian noise channel.
In these solutions, the very low contrast image ($\lambda_{\text{Max}} = 20$) shows slightly better restoration result than that of ours, however, the larger contrast becomes, the lower the performance becomes.
From the eq.(\ref{eq:likelihood2}), 
our method could express the observation accuracy $\beta_i$ for each pixel value, 
however, 
the GLBP solution has only single hyperparameter $\beta_G$ in eq.(\ref{eq:Gauss1}).
The BM3D also assumes the variance of the pixels in a image might be denoted 
as a single parameter in its algorithm.
The variance and the mean of the Poisson observation, however, 
depends on the single parameter $\lambda_i$, so that, the assumption of 
GLBP and BM3D observation might not satisfy.
As the result, in the GLBP case, the $\beta_G$ tends to be overestimation.
In the numerical evaluation, our method shows better performance rather than that 
of the GLBP except $\lambda_{\text{Max}} = 20$.
Thus, our results suggests that considering the correct Poisson observation model 
is important as well as the choosing of the prior.

Our latent variable method evaluate the Poisson likelihood function
as the Gaussian form. 
Thus, in future works, 
we can easily extend our method into other image restoration framework.
For example, when we could express the the parameter values of $\Vec{x}$ as
the some linear transformation:
\begin{align}
  \Vec{x} = A \Vec{s}
\end{align}
where $A$ and $\Vec{s}$ are the transformation matrix and 
the expression vector respectively.
Then, we could substitute it into the eq.(\ref{eq:likelihood1}) and consider the
prior for the transformed vector $\Vec{s}$ such like sparse prior,
such like K-SVD\cite{Ahron06}.

\section*{Acknowledgments}
We appreciate K.~Takiyama, Tamagawa-Gakuen University, and
Professor M.~Okada, University of Tokyo for fruitful discussions.
This work is partly supported by MEXT/JSPS KAKENHI Grant number 25330285, and 26120515.

\bibliographystyle{unsrt}
\bibliography{shouno}

%\begin{biography}
%\profile{Shouno Hayaru}{was born in 1968. He received M.E.\ and
%Ph.D.\ from Osaka University in 1994 and 1999. His current research interest 
%is image processing and artificial neural network.
%He is a member of the IEEE, the IEICE, and IPSJ.
%}
%\end{biography}

\end{document}